\documentclass[a4paper,11pt]{article}

\usepackage{geometry}
\usepackage[utf8]{inputenc}
\DeclareUnicodeCharacter{2212}{-}
\usepackage{amsmath}
\usepackage{microtype}
\usepackage{siunitx}
\usepackage{authblk}
\usepackage{lscape}
\usepackage{graphicx}
\usepackage{subcaption}
\usepackage[font={footnotesize},labelfont=bf]{caption}
\usepackage[super,numbers,sort&compress]{natbib}
\usepackage{booktabs}
\usepackage{blindtext}
\usepackage{booktabs,multirow,array}
\usepackage{tikz}
\usepackage{verbatim}
\usepackage{chemfig}
\usepackage{gensymb}
\usepackage{textcomp}
\usepackage{setspace}
\usepackage{multicol}
\usepackage{filecontents,pgfplots,pgfplotstable}
\usepackage{afterpage}
\usepackage{xtab,afterpage}
\usepackage{amssymb}
\usepackage{rotating}
\usepackage{natbib}
\usepackage{scrextend}
\usepackage{tcolorbox}
\usepackage[hidelinks]{hyperref}
\usepackage{cleveref}
\usepackage{algorithm}
\usepackage[noend]{algpseudocode}
\usepackage{enumitem}
\setlist[itemize]{leftmargin=*}
\usepackage{microtype}
\usepackage{float}

\usepackage{arxiv}

\begin{document}

\title{Generating 3D structures from a 2D slice with GAN-based
dimensionality expansion}

\author[1]{{Steve Kench}
\thanks{s.kench19@imperial.ac.uk}\ \ }

\author[1]{Samuel J. Cooper \thanks{samuel.cooper@imperial.ac.uk}\ \ }

\affil[1]{{\textit{\footnotesize Dyson School of Design Engineering, Imperial College London, London SW7 2DB}}}

\maketitle

\begin{abstract}
\begin{center}
\begin{minipage}{0.85\textwidth}
{\small Generative adversarial networks (GANs) can be trained to generate 3D image data, which is useful for design optimisation. However, this conventionally requires 3D training data, which is challenging to obtain. 2D imaging techniques tend to be faster, higher resolution, better at phase identification and more widely available. Here, we introduce a generative adversarial network architecture, SliceGAN, which is able to synthesise high fidelity 3D datasets using a single representative 2D image. This is especially relevant for the task of material microstructure generation, as a cross-sectional micrograph can contain sufficient information to statistically reconstruct 3D samples. Our architecture implements the concept of uniform information density, which both ensures that generated volumes are equally high quality at all points in space, and that arbitrarily large volumes can be generated. SliceGAN has been successfully trained on a diverse set of materials, demonstrating the widespread applicability of this tool. The quality of generated micrographs is shown through a statistical comparison of synthetic and real datasets of a battery electrode in terms of key microstructural metrics. Finally, we find that the generation time for a $10^8$ voxel volume is on the order of a few seconds, yielding a path for future studies into high-throughput microstructural optimisation.}
\end{minipage}
\end{center}
\end{abstract}
\vspace{.2cm}
\begin{multicols}{2}
\section{Introduction}
\label{Introduction}
 The properties and behaviour of a material depend strongly on its microstructure, which in turn is controlled by conditions of manufacture. Physical simulations play an important role in offering insight into these relationships to help inform the design of next generation materials. Importantly, many material behaviours, such as deformation under stress or fluid flow through a porous medium, are inherently volumetric and cannot be accurately modelled using 2D data alone. The fidelity of simulation techniques used to extract these properties is thus partly determined by the quality of available 3D microstructural datasets. However, as compared to their 3D equivalents, 2D micrographs are often easier to obtain, higher resolution and can contain a more representative distribution of features based on a larger field of view. Accurate methods for the statistical reconstruction of 3D volumes using 2D micrographs are thus highly desirable.

Generative adversarial networks (GANs) are a promising candidate model for this task. A GAN consists of two neural nets: a generator, $G$, to synthesise fake samples, $\textbf{f}$, and a discriminator, $D$, to distinguish between $ \textbf{f}  $ and real samples, , $ \textbf{r}  $, from the dataset. During training, $G$ and $D$ are updated iteratively, enabling the generator to capture features of the real dataset and ultimately, if successful, produce realistic images \cite{goodfellow2014generative,creswell2018generative,odena2016semi}. Since their introduction, extensive work has led to improved image quality, feature diversity and training stability \cite{arjovsky2017wasserstein}. The motivation behind this paper is to apply this powerful technique to the task of dimensionality expansion. Crucially, in order to achieve this, the incompatibility between a 3D generator versus 2D training data must be solved. Furthermore, important considerations regarding the effect of information density on image quality are raised. The key contributions of this work are summarised as follows:

\begin{enumerate}[wide, labelwidth=!, labelindent=0pt]
	\item In section \ref{SliceGAN}, we introduce a novel GAN architecture, \emph{SliceGAN}, which is designed to address the challenge of generating 3D volumetric data from 2D slices of an isotropic material. The ability to statistically reconstruct anisotropic microstructures with a simple extension is also demonstrated.
	  
	\item In section \ref{info}, the low quality regions found at the outer edges of GAN generated images is discussed. This issue is found to be associated with non-uniform information density in the generator. The authors define a set of requirements for the parameters of transpose convolutional operations to avoid this issue.
	
	\item In section \ref{Results}, \emph{SliceGAN} is applied to a range of 2D microstructures, and shown to produce 3D volumes with visually indistinguishable cross sections from training data. The seven example training sets include a synthetic crystalline microstructure, an anisotropic polymer membranes and a three phase battery electrode material. This final example is further validated through a comparison of three key metrics to demonstrate the statistical similarity between synthetic volumes and the original dataset.
	
\end{enumerate}

\section{Background}

\subsection{GANs and their application in material science}
Deep convolutional generative adversarial networks use a layered architecture to identify and reproduce the hierarchical features of an image dataset. When trained successfully, they offer high resolution, diverse image generation. Since the introduction of GANs, numerous adaptions have been proposed,  falling broadly into two categories. First, there are those dedicated to improved training of adversarial nets, such as Wasserstein GANs (WGANs), which increase training stability and generated image diversity by addressing the challenge of vanishing gradients \cite{arjovsky2017wasserstein,Arjovskyimproved}. Second, there are approaches that aim to bring new capabilities such as conditional GANs (cGANs), which are used to condition a generator based on image labels, thus allowing the synthesis of multiple image categories using a single generator \cite{mirza2014conditional}. A further relevant example of this latter category developed a method for spatial model generation from a set of 3D training samples \cite{Wu}. The diverse and rapid evolution of these different methodologies indicates the flexibility and wide applicability of GANs.\\

It is thus unsurprising that GANs have already proven useful in the field of material science \cite{Mosser}. A recent study by Gayon-Lombardo et al. used GANs to generate the microstructure of energy storage materials based on 3D training data \cite{gayonlombardo2020pores}. The synthetic volumes were not only much larger than the training set, but also had periodic boundaries, which makes them particularly well suited to generating representative domains for multi-physics simulation. GANs have also been applied to the task of material design. Li $ et.$ $ al $ trained a generator to synthesise realistic micrographs of a semi-conductor surface \cite{yang2018microstructural}. A batch of instances can then be produced by the generator, and the property of interest, in this case absorption co-efficient, calculated for each sample. A Gaussian processes algorithm was then used to optimise for this property through an informed exploration of the microstructural space. This offers an efficient alternative compared to a conventional experimental materials design approach. These studies are an indication of the broad potential applications of GANs in the field of material science.

\subsection{Current microstructural reconstruction approaches}
Due to the widespread potential applications of 2D to 3D reconstruction algorithms, a range of techniques have been developed in previous studies. For example, given a particle size distribution measured from a 2D image, simulated annealing algorithms can be used alongside ellipsoid packing to emulate powder processing \cite{xu2014descriptor}. Equally, tessellation algorithms are incorporated in software tools such as Dream3D, which uses extracted 2D properties such as grain size and orientation to simulate 3D grain growth \cite{groeber2014dream}. Unfortunately, these techniques are designed for, and thus limited to, powder and polycrystalline microstructures respectively. Furthermore, a loss of geometric information is inevitable, as irregular particles are approximated to ellipsoids, and complex grain shapes are characterised by mean diameters and aspect ratios.\\

\begin{figure*}[hb]
\centering
\includegraphics[width=\textwidth]{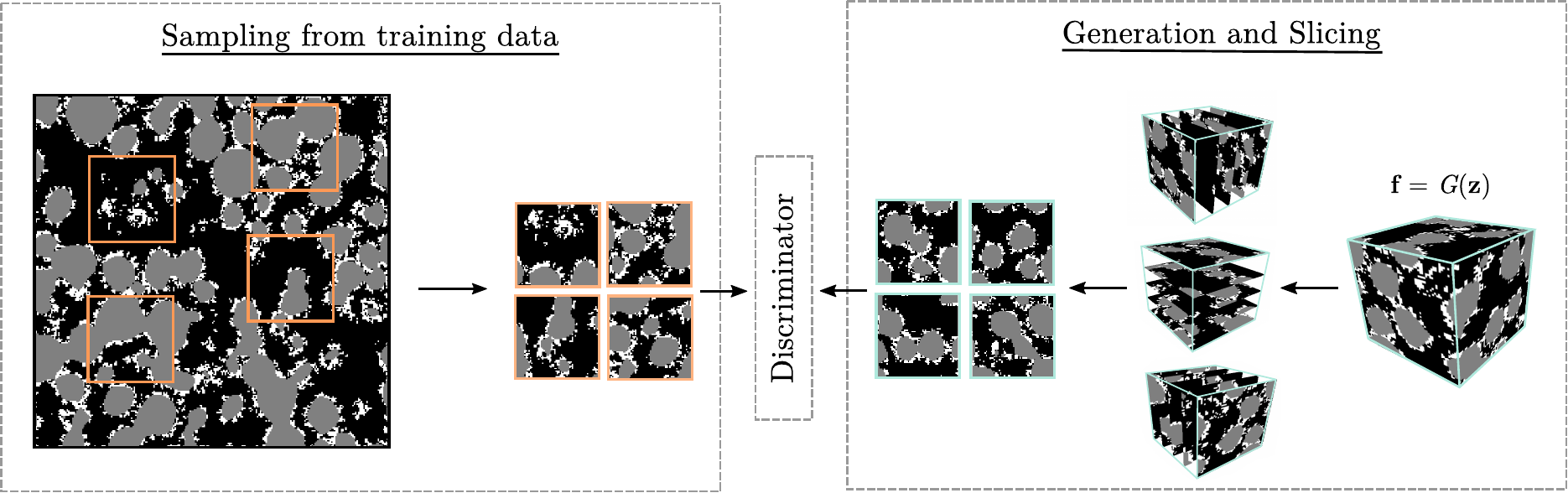}
\caption{{\footnotesize \emph{SliceGAN} training procedure. First, real images are sampled from a 2D training sample. Second, a fake volume $\textbf{f}$ is generated and sliced along $ x $, $ y $ and $ z $. This yields a compatible pair of datasets, which can both be fed to a 2D discriminator.}}
\label{fig:SliceGANfig}
\end{figure*}\vspace{-0.2cm}

Other approaches have been developed based on correlation statistics. Given an image containing $ i $ phases, the two-point correlation function $ S_2 $ describes the probability that two randomly selected pixels, separated by distance $ r $, are of the same phase \cite{torquato1982microstructure}. Importantly, it has been shown analytically that for a given isotropic material, the same two-point correlation function is recovered whether calculated for a 2D or 3D dataset\cite{torquato2002random}. This led to the development of stochastic algorithms which take initially random 3D volumes and iteratively attempt to minimise the difference between the correlation functions of the real and generated data \cite{hasanabadi20163d,izadi2017application}. Whilst successful in some cases, the degeneracy of two-point correlations can lead to unrealistic results \cite{gommes2012microstructural}. Furthermore, even state-of-the-art methods are computationally expensive with simulation times to synthesise a modest $10^6$ voxel sample on the order of hours using a 2.8 GHz, 16GB RAM computer, making this approach unsuitable for materials optimisation approaches \cite{zhang2019high}. \\

The GANs described in this work take a similar time span to train (c. 4 hours on an NVIDIA Titan Xp GPU); however, once trained, the generator is able to synthesis $10^8$ voxel instances in seconds. This $10^5$ acceleration in generation speed compared to \cite{zhang2019high} enables both high-throughput optimisation and large scale modelling. Furthermore, GANs are highly flexible compared to the tessellation algorithms described above. Indeed, a key strength of deep convolutional neural nets is their ability to efficiently capture the diverse and complex features of an image without any user defined statistical metrics as inputs. Another important advantage of using GANs is the breadth of available variants that could be incorporated. Adaptations might aim to decrease training time, for example through transfer learning, or to facilitate new capabilities, for example using conditional GANs to allow the generation of different microstructures with specific features indicated by a label. \emph{SliceGAN} is thus a building block for a powerful tool set of machine learning methods for microstructural characterisation, modelling and optimisation.

\section{SliceGAN}
\label{SliceGAN}

The fundamental role of \emph{SliceGAN} is to resolve the dimensionality incompatibility between a 2D training image and 3D generated volumes. This is achieved by incorporating a slicing step before fake instances from the 3D generator are sent to the 2D discriminator. For a generated cubic volume of edge length $l$ voxels, $3l$ 2D images are obtained by taking slices along the $x, y $ and $z$ directions at 1 voxel increments. During the training of $D$, for each fake 2D slice, a real 2D image is also sampled and fed to the discriminator, such that it learns equally from real and fake instances. To train $G$, the same slicing procedure is followed; however, a larger batch size ($m_G$) compared to the discriminator ($m_D$) is used. This rebalances the effect of training $D$ on a large number of slices per generated sample. We find that $m_G = 2 m_D$ typically results in the best efficiency. Importantly, the particular architecture of $ G $ used in this study means that each slice in any given set of 32 adjacent planes is synthesised through a unique combination of kernel elements (see section \ref{info}). As such, a minimum of 32 slices in each direction must be shown to $ D $ in order to ensure each path through the generator is trained. In practice, we find training to be both more reliable and efficient when $ D $ is applied to all 64 slices in each direction, despite the doubling of discriminator operations that this entails. After slicing, a standard Wasserstein loss function is used as proposed in \cite{arjovsky2017wasserstein} to encourage stable training. The complete training process for \emph{SliceGAN} is described in Algorithm \ref{alg:SliceGanAlgo}, with the data sampling, generation and slicing stages depicted in Figure \ref{fig:SliceGANfig}.\\

\begin{algorithm*}[ht!] 
	\begin{algorithmic}[1]
		\Require $w$ and $\theta$, the trainable parameters for $D$ and $G$ respectively; the number of discriminator iterations per generator iteration $n_D$; the batch sizes $m_D$ and $m_G$ for $D$ and $G$ respectively; the gradient penalty coefficient, $\lambda$;  volume edge length, $l$; and Adam optimizer hyperparameters \cite{kingma2017adam}, $\alpha, \beta1, \beta2 $.
		\While {$ \theta $ has not converged }
		\State Discriminator training:
		\For {$t=0,...,n_{{D}}$ }
		\For {$i = 1, ..., m_D$}
		\State Sample a latent vector from a normal distribution $ \textbf{z} \sim p(\textbf{z})  $ 
		\State $\textbf{f} \leftarrow G_{\theta}(\textbf{z})$ generate a 3D volume
		\For {$a = 1,2,3$}
		\For {$d = 1,...,l$}
		\State $\textbf{f}_{\textbf{s}} \leftarrow$  2D slice of $\textbf{f}$ at depth $d$ along axis $a$
		\State Sample an $l \times l $ image from the real dataset $ \textbf{r} \sim \mathbf{P}_\textbf{r}$ 
		\State Sample a random number $\epsilon \sim \mathbb{U}[0,1]$
		\State $\textbf{k} \leftarrow \epsilon \textbf{f}_{\textbf{s}} + (1-\epsilon)\textbf{r}$
		\State $L_D^{(a,d)} \leftarrow D_w(\textbf{f}_\textbf{s}) - D_w(\textbf{r}) + \lambda(\Vert \nabla_{\textbf{k}} D_w(\textbf{k}) \Vert_2-1)^2$
		\EndFor
		\EndFor
		\State $w \leftarrow \text{Adam}(\nabla_w \frac{1}{m_D}\sum_{a = 1}^{3}\sum_{d = 1}^{l}
		 L_D^{(a,d)})$
		\EndFor
		\EndFor
		\State Generator Training:
		\For {$j = 1,...,m_G$}
		\State Sample a latent vector from a normal distribution $ \textbf{z} \sim p(\textbf{z})  $ 
		\State $\textbf{f} \leftarrow G_{\theta}(\textbf{z})$ generate a 3D volume
		\For {$a = 1,2,3$}
		\For {$d = 1,...,l$}
		\State $\textbf{f}_{\textbf{s}} \leftarrow$  2D slice of $\textbf{f}$ at depth $d$ along axis $a$
		\State $L_G^{(a,d)} \leftarrow -D_w(\textbf{f}_{\textbf{s}})$
		\EndFor
		\EndFor
		\State $\theta \leftarrow \text{Adam}(\nabla_\theta \frac{1}{m_G}\sum_{a = 1}^{3}\sum_{d = 1}^{l} L_G^{(a,d)})$
		\EndFor

		\EndWhile

		\caption{{\footnotesize SliceGAN algorithm for isotropic materials}}
		\label{alg:SliceGanAlgo}
	\end{algorithmic}
\end{algorithm*}\vspace{-0.5cm}

Importantly, the above approach is only suitable for isotropic materials, where a single 2D image can be used to train slices from the $x, y$ and $z$ orientations of $\textbf{f}$. An extension of the architecture is required for anisotropic materials. For example, given a microstructure containing a matrix with rods oriented parallel to the $z$ axis, two training images are needed: one perpendicular to the $z$ axis containing circular features and another parallel to the $ z  $ axis containing long rectangular features (whilst these are the simplest orientations for this example, any set of perpendicular planes could be chosen in principal). During training, images $\textbf{r}_{z}$ are sampled from the former and images $\textbf{r}_{x,y}$ from the latter, whilst slices from the fake samples are also divided into those taken along the $z$ axis, $\textbf{f}_{z}$, and those taken from $ x $ or $y$, $\textbf{f}_{x,y}$. Separate discriminators are then taught to capture the distribution of features along the different orientations, and thus teach $G$ to synthesise volumes that look like $\textbf{r}_{z}$ along the $z$ axis, and $\textbf{r}_{x,y}$ along the $ x $ and $ y $ axis. The adjusted \emph{SliceGAN} algorithm for anisotropic materials is given in Supplementary Information S1. Successful examples of this approach are presented in section \ref{Results}.

\section{Generator information density} \label{info}

\begin{figure*}[ht!]
	\includegraphics[width=\textwidth,]{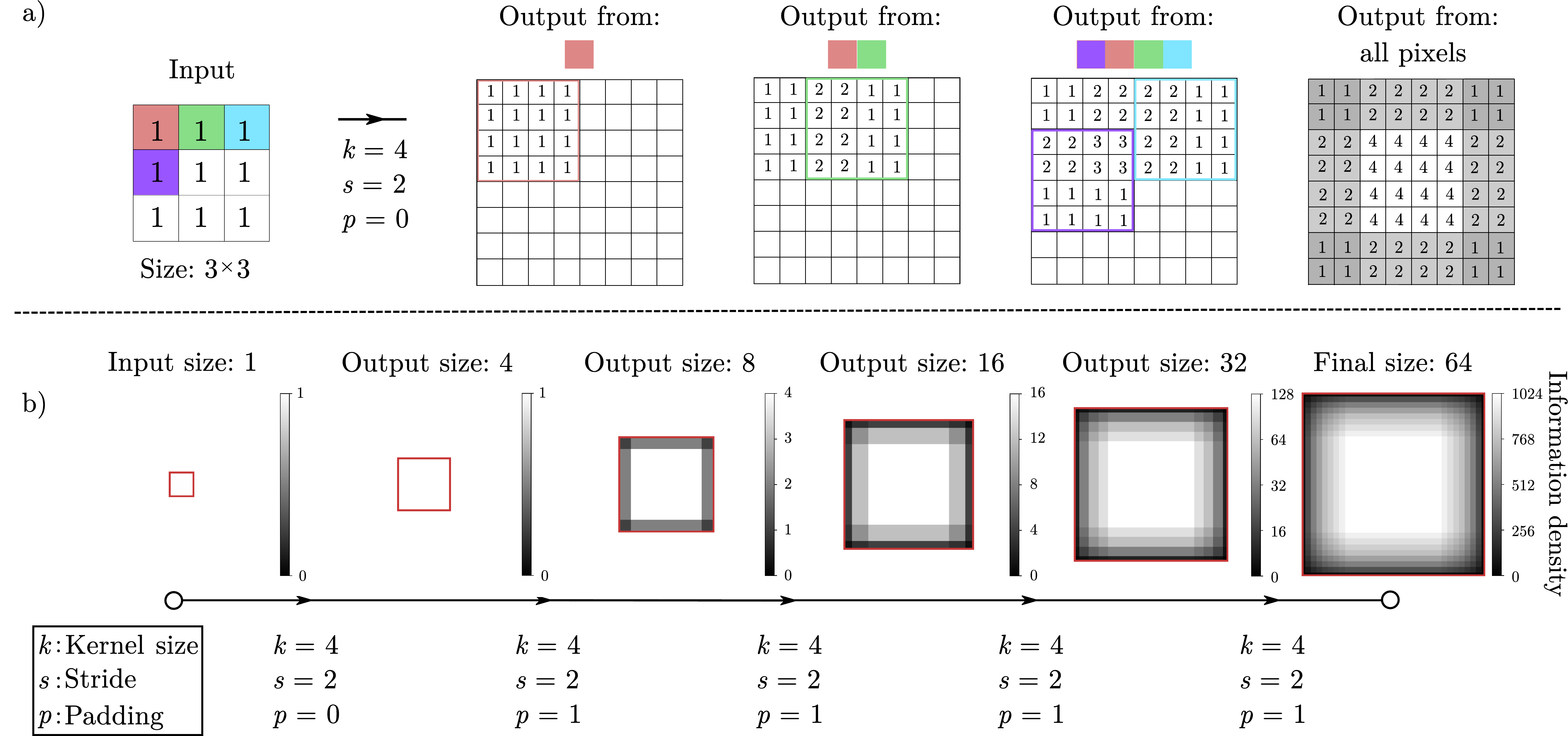}
	\caption{{\footnotesize a) An example of how information density is calculated for a single transpose convolution with input size = $3 \times 3$,  $k$ = 4, $ s $ = 2 and $p$ = 0. The overlap from the outputs of individual pixels is depicted, showing a information density gradient at the edges of the final image. b) Illustration of how information density gradients propagate and are exacerbated in a full neural net as an input is passed through multiple transpose convolutional layers.}}
	\label{infodensity}
\end{figure*}

During the development stages of \emph{SliceGAN}, significantly lower visual quality was observed near sample edges compared to the bulk. This may not be problematic for GANs applications such as face generation, where key objects often lie in the centre of an image and the edges are of less importance. However, for microstructural synthesis this is not the case, as although the relative position of features is important, their absolute location is often arbitrary. Edges are thus of equal importance to the center, and can strongly influence the measured physical properties of a microstructure. To understand the cause of edge artifacts, a simple test case is proposed to explore the nature of transpose convolutions, which are the key operator within the convolutional generator neural nets used in this work. We first define a $3 \times 3$ input with a single channel, and pixel values set to 1. The transpose convolution parameters are selected with kernel size of $ k = 4$; stride, $ s =2$; padding,  $ p = 0$ (NB. In a generator, $ p $ refers to the removal of near edge layers, rather than their addition) and output channels $C_{\text{out}} = 1$. The kernel weights are set to 1. Figure \ref{infodensity}a depicts this simple scenario, and shows how the transpose convolution operates on the input pixel by pixel. We first observe the output from the top left pixel, which is distributed over a region equal to the kernel size. As the contribution from subsequent pixels is added, an overlap of information is seen. The final output shows a central region with flat information density of 4, and lower information content towards the edges and corners. Figure \ref{infodensity}b shows the additive effect of this phenomenon in a typical generator with multiple convolutional operations. Indeed, the number of parameters contributing to central pixels is 3 orders of magnitude greater than for edge pixels, which explains the poor image quality in this region.\\

 In order to avoid artifacts of this nature, three rules can be used to define acceptable combinations of $ k, s $ and $ p $ such that a uniform information density is ensured (Although an important consideration for image quality, $C_{\text{out}}$ does not effect information density gradients):
\begin{enumerate}[wide, labelwidth=!, labelindent=2pt]
	\item\  $ s < k $ (Although $s \leq k $ is a sufficient requirement for uniformity, $ s < k $ ensures that there is kernel overlap which is crucial for avoiding feature discontinuities in the output.)
	\item\  $ k \bmod s = 0$ (If $k$ is not divisible by $s$, the kernel output regions from nearby input pixels never lie flush to each other, resulting in a checkerboard information density \cite{odena2016deconvolution}. An example of this behaviour with $k = 4$, $ s= 3$ is shown in Supplementary Information S2.)
	\item\  $ p\geq k-s $ (The region of edge voxels with less than the central uniform information density is of width $k-s$, and must thus be removed by applying a padding of this value or greater.)
\end{enumerate}

These constraints leave a small set of practical transpose convolutions. We first discard any prime value of $k$, as in this scenario rule 1 and 2 together only allow for $s=1$. Such configurations result in no expansion of the spatial dimensions and are thus not useful. In general, $k>6$ is also impractical, especially in 3D convolutional nets, as the number of parameters scales with $k^3$. Assuming $p$ is chosen to be as small as possible to minimise the amount of cropping at each layer (which equates to wasted computational expense), this leaves three practical sets for $\{k,s,p\}$: $\{4,2,2\}, \{6,3,3\},\{6,2,4\}$.  In the work presented here, the $\{4,2,2\}$ set of parameters are used for most transpose convolutions to avoid the computational expense of $k=6$. An alternative approach to avoid edge artifacts could be the use of resize-convolution generator architectures with a different set of parameter rules (most importantly, the use of any zero-padding results in undesirable information gradients). Unfortunately, under these constraints resize-convolution has significantly larger memory requirements for 3D volume generation than transpose convolutions. A sacrifice in either training time (by reducing batch size) or image quality (by reducing the number of filters) thus becomes necessary, as shown in Supplementary Information S3. \\

\begin{table*}[ht!] 
    \vspace{-0.2cm}
	\centering
	\caption{{\footnotesize Table showing the architecture of the 3D generator (memory size $\approx$ 50 MB) and 2D discriminator (memory size $\approx$ 11 MB) used in this work.} \label{Tab:ArchTab}}
	\begin{tabular}{ |c|c|c|c|c|c|} 
		\hline
		\multicolumn{3}{|c|}{Generator}   & \multicolumn{3}{|c|}{Discriminator}  \\ 
		\hline
		Layer & $ k, s, p  $ &Output shape &  Layer &  $ k, s, p  $& Output shape \\ 
		\hline
		Input $\mathbf{z}$ &  - & $64 \times 4 \times 4 \times 4$ &Input &  - & $3 \times 64 \times 64 $\\
		1 &  4,2,2& $512 \times 6 \times 6 \times 6$ &1&  4,2,1& $64 \times 32 \times 32$ \\
		2 &  4,2,2& $256 \times 10 \times 10 \times 10$& 2 &  4,2,1& $128 \times 16  \times 16$\\
		3 &  4,2,2& $128 \times 18 \times 18 \times 18$ &3 &  4,2,1& $256\times 8 \times 8$\\
		4 &  4,2,2& $64 \times 34 \times 34 \times 34$ &4&  4,2,1& $512\times 4 \times 4$\\
		5 &  4,2,3& $3 \times 64 \times 64 \times 64$& 5 &  4,2,0& $1 \times 1 \times 1$\\	
		softmax &  -&$3 \times 64 \times 64 \times 64$&&& \\
		\hline
	\end{tabular}
	\vspace{0.1cm}
\end{table*}

    \begin{table*}[hb!]
  	\centering
  	  	\caption{{\footnotesize Synthesis details for the examples presented in Figure \ref{fig:Resultsfig}.}}
  	\begin{tabular}{ |c|c|c|c|c|c|} 
  		
  		\hline
  		Label & Material & Imaging technique & Image type & Isotropy & Ref \\ 
  		\hline
  		A & Polycrystalline grains & Synthetic & Two-phase& Isotropic & -\\
  		B & Ceramic (perovskite) & Kelvin probe force topography & Two-phase& Isotropic& \cite{gratia2017many}\\
  		C & Carbon fibre rods & Secondary electron microscopy & Two-phase& Anisotropic&  \cite{LeeArmentrout}\\
  		D& Battery seperator & X-ray tomography reconstruction & Three-phase& Anisotropic & \cite{Finegan2016} \\
  		E & Steel & Electron back-scatter microscopy & Colour& Isotropic &\cite{OxInst}\\
  		F & Grain boundary  & Synthetic & Three-phase& Anisotropic & - \\
  		G & NMC battery cathode & X-ray tomography & Three-phase& Isotropic & \cite{hsu2018mesoscale}\\
  		\hline
  	\end{tabular}
	\vspace{0.3cm}
  	\label{Tab:EmpDetails}
  \end{table*}
  
Another crucial attribute when designing a GAN architecture for material science applications is the ability to generate different size volumes during inference. This is important as often the standard training volume (normally 64 voxels cubed to allow reasonably large batch sizes under memory constraints) is not large enough for use in multiphysics simulations. A previous study suggests that after training, bespoke volumes can be synthesised by varying the spatial size of the latent vector $ \mathbf{z}$ \cite{gayonlombardo2020pores}. However, this approach results in lower quality microstructures with distortions. This observation can be explained by the fact that during training, $\mathbf{z}$ has a spatial dimension of 1, making the first transpose convolution of $G$ unique in that it operates on a single voxel per channel. In this scenario, no kernel overlap occurs in the first layer output. Increasing the spatial dimension of $\mathbf{z}$  during inference changes this behaviour and introduces overlap. To avoid this problem, we choose to train into the first generator layer an understanding of overlap; thus an input vector with spatial size 4 is used. The resulting information spread is depicted in Supplementary Information S4, with the final network architectures given in Table \ref{Tab:ArchTab}. Interestingly, this alteration leads to periodicity within the generator, whereby sets of every 32$^{\text{nd}}$ plane are generated from the same combination of kernel elements, as shown in Supplementary Information S5. This does not make the synthetic volume itself periodic, as each plane is associated with a different region of the input seed. However, it does mean that it is not possible to train $G$ to capture anisotropy within a given plane (e.g a graded density).

\section{Empirical results}\label{Results}

\subsection{Data pre-processing}
Whilst it is possible to train \emph{SliceGAN} using RGB or greyscale micrographs (for which no pre-processing is necessary), the approach has been most successful with one-hot encoded representations of segmented $ n $-phase microstructural data. For a training dataset, the encoding process entails the creation of $ n $ separate channels, each containing 1s where the material phase they correspond to is present and 0s otherwise, as shown in Supplementary Information S6. In generated instances, $ n $ channels are again required, though they now represent the probability of finding a given phase at a location in an image. This is implemented using a \textit{softmax} function as the final layer of the generator. In either case, an $n$-phase microstructure can always be recovered from a one-hot array through a \textit{max} function over the channel axis.

\subsection{Application to microstructures}

  \begin{figure*}[ht!]
  	\centering
  	\includegraphics[width=\textwidth]{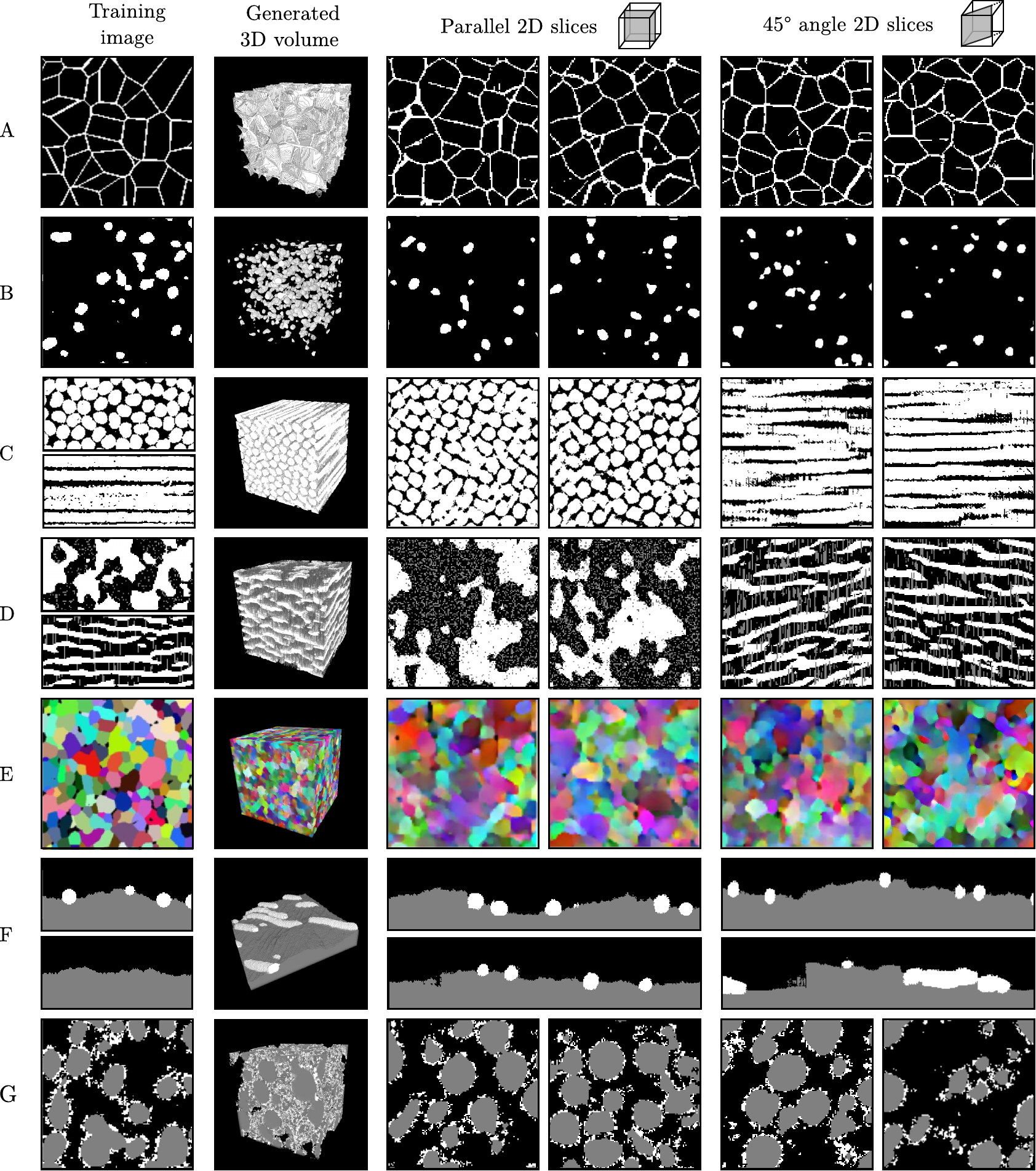}
  	\caption[]
	{{\footnotesize Application of \emph{SliceGAN} to a variety of microstructures, with details specified in Table \ref{Tab:EmpDetails}. From left to right, the original training dataset, a 3D statistical reconstruction and 2 examples of slices through an example volume taken at different angles.}}
  	\label{fig:Resultsfig}
  \end{figure*}
  
To demonstrate the capabilities of \emph{SliceGAN}, a range of micrographs were chosen to be statistically reconstructed, as shown in Figure \ref{fig:Resultsfig}. The first column shows a region of the 2D training data used. The second column shows an example of a generated 3D instance, with the volume edge lengths selected to best display the microstructural features. Next, slices of a synthetic volume taken i) parallel to a discriminated plane to show the dataset quality and ii) at a 45$^{\circ}$ angle to the discriminators to demonstrate the consistency of the microstructure in directions not involved in training. For each case, the material, imaging technique and training details are given in Table \ref{Tab:EmpDetails}. The first example presented in row A of Figure \ref{fig:Resultsfig} is a synthetic grain structure, chosen for its well defined features: perfectly straight grain boundaries and exclusively convex grains. \emph{SliceGAN} performs reasonably well on this data, generating realistic looking volumes, though some disconnected and curved grain boundaries are observed. These shortcomings are not specific to the \emph{SliceGAN} approach, as similar features are also observed in synthetic 2D images from a conventional GAN. The next training image (row B) is a real micrograph from surface analysis of a ceramic, where the black phase correspond to areas with high Caesium content \cite{gratia2017many}. Again, a visually realistic sample is generated through \emph{SliceGAN}.\\

  \begin{figure*}[ht!]
	\centering
	\includegraphics[width=\textwidth,]{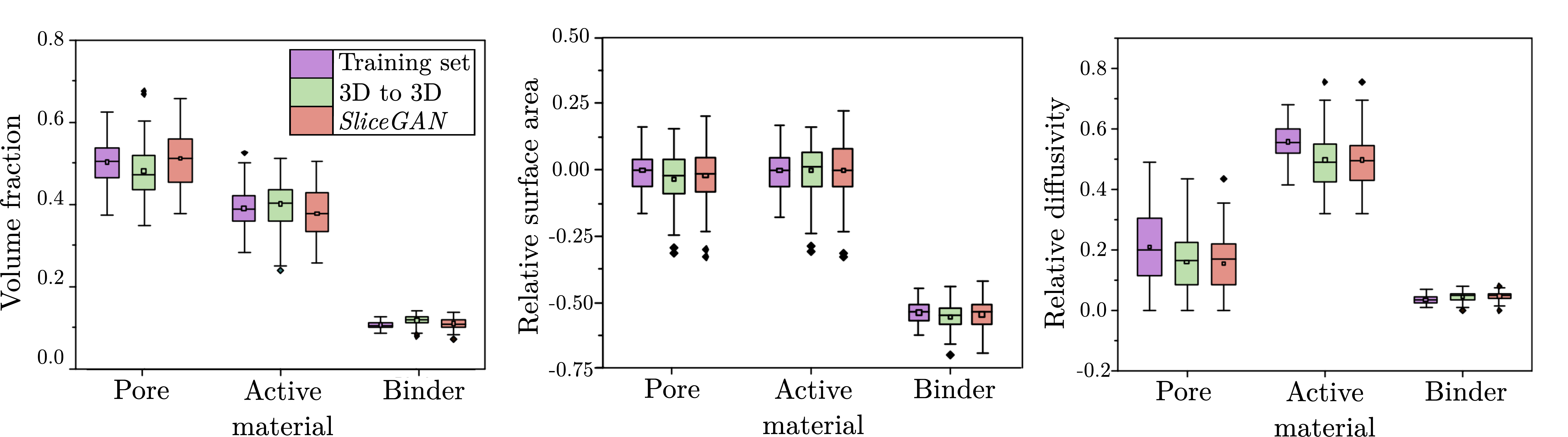}
	\caption{{\footnotesize Statistical analysis of three electrochemical and general materials properties, showing that the real properties of the dataset are captured well by both a 3D to 3D GAN and \emph{SliceGAN}. Boxes depict the interquartile range (IQR) from the lower quartile, Q1 to the upper quartile, Q3). Each box contains a horizontal line and a small square to indicate the median and mean respectively. The upper and lower whiskers show the last datum greater than $\text{Q3} + (1.5 \times \text{IQR})$ and less than $\text{Q1} - (1.5 \times \text{IQR})$ respectively. Outlier data points beyond this range are represented by solid black diamonds. }}
	\label{fig:StatAnal}
\end{figure*}
  
Two images of fibre reinforced composite taken from perpendicular views demonstrate the ability to synthesise anisotropic images \cite{LeeArmentrout}. A similar approach is taken for a polymer battery separator material \cite{Finegan2016,Xu2020}. This demonstrates the flexibility of \emph{SliceGAN}, and will enable its application to many common microstructures (rows C and D). A micrograph of polycrystalline single phase steel is considered next (row E) \cite{OxInst}. Previous work was focused on segmented 2 or 3 phase materials; here the extension to colour images (where color encodes grain orientation) is presented. Image quality is somewhat reduced compared to previous examples. This is likely because our 12 GB GPU memory constraint means that the networks are too small to capture continuous label spaces (instead of the segmented materials in other examples). These volumes can be made to look more like the training data by applying filters (e.g. mode), but this kind of post-processing is beyond the scope of this study and would need to be selected on a case by case based. In row F, a test on a synthetic material interface is presented where a large 128 $\times 10^4$  pixel image was generated using a bounded random walk algorithm to generate an interface, followed by seeding of randomly size boundary particles. In this case, the 2D example images shown are small crops of the full training dataset. This was an opportunity to test whether 2 discriminators, both perpendicular to the material interface, were sufficient for training. Interestingly, whilst the general surface topology of the synthesised material reflects the real dataset well, the interface particles form long ridges on the surface at a 45$^{\circ}$ angle. This is due to insufficient constraint without the top view of the material, which allows the generator to effectively trick the discriminator. This could potentially be resolved by implementing discriminated slices at 45$^{\circ}$ angles.\\

  Finally, open access tomographic data collected from a Li-ion NMC cathode sample is used as shown in row G \cite{hsu2018mesoscale}. Although the dataset is 3D, only a random subsample of 2D sections are used for \emph{SliceGAN} training. This experiment is particularly useful as 3D properties of the original dataset can be compared to those of the synthesised microstructure. Furthermore, instances from a 3D to 3D GAN can also be evaluated. Thus, to quantify the performance of \emph{SliceGAN}, 100 sample batches were taken from each of i) the real dataset ii) the 3D to 3D dataset generated by Gayon-Lombardo et al \cite{gayonlombardo2020pores} and iii) \emph{SliceGAN}. Three relevant metrics were chosen and calculated using \emph{TauFactor} \cite{COOPER2016203} (an open-source microstructural analysis package) for each set of microstructures, as shown in Figure \ref{fig:StatAnal}.  Two-point correlation functions and triple phase boundary densities are also compared in Supplementary Information S7.

  For all of these conventional microstructural metrics, both GAN architectures reproduce the training set well. Although the medians correspond very closely in each case, the distributions, in particular in the case of the relative diffusivity, are not exactly the same as the training set. Crucially, unlike the other metrics, the effective diffusivity is an emergent property of the generated 3D volume as a whole, and is not measurable from any individual 2D slice. As such, it’s particularly impressive that \emph{SliceGAN} can reproduce this metric.

\section{Conclusions}
The results presented in this study show the ability of \emph{SliceGAN} to faithfully synthesise 3D $ n $-phase media from 2D micrographs. This tool has three key uses: First, in the field of material characterisation as a means to visualise a statistically realistic 3D instance of a 2D micrograph. This is especially important considering the limitations of direct 3D imaging, namely in terms of resolution and field of view. Second, for electrochemical simulation; 2D micrographs are insufficient for a range of desirable electrochemical and mechanical simulation techniques. Finally, for materials optimisation purposes, as the speed of the generator allows rapid exploration of the space of possible 3D microstructures. Future work will aim to develop this tool set in combination with other common GANs approaches such as conditional GANs and transfer learning. The former will enable interpolation between microstructures with differing properties, whilst the latter aims to speed the training process. Thus, there is significant potential for expansion of this approach as a tool for material characterisation and optimisation.

\section*{Data Availability}
The study used open-access training data available from the following sources: ceramic \cite{gratia2017many}, carbon fibre rods \cite{LeeArmentrout}, battery separator \cite{Finegan2016,Xu2020}, steel \cite{OxInst}, and NMC battery cathode \cite{hsu2018mesoscale}. 

\section*{Code Availability}
 The codes used in this manuscript, are available at: \url{https://github.com/stke9/SliceGAN}. All other generated data used is available from the authors on request.

Correspondence and requests for materials should be directed to Steven Kench at sk3619@ic.ac.uk.

\section*{Acknowledgements}

This work was supported by funding from the EPSRC Faraday Institution Multi-Scale Modelling project (https://faraday.ac.uk/; EP/S003053/1, grant number FIRG003 received by SK). The Titan Xp GPU used for this research was kindly donated by the NVIDIA Corporation through their GPU Grant program to SC. The authors would also like to thank Aron Walsh, Andrea Gayon Lombardo and Daniel Figueroa for their valuable comments, as well at the rest of the tldr group.

\section*{Authorship}

SK designed and developed the code for $ SliceGAN $, trained the models, performed the statistical analysis and drafted the manuscript. SJC contributed to the development of the concepts presented in all sections of this work, helped with data interpretation in section 5 and made substantial revisions and edits to all sections of the draft manuscript.

\section*{Competing interests}
The authors declare no competing interests.

\section*{References}
\addcontentsline{toc}{section}{References}

\def\addvspace#1{}

	\renewcommand{\refname}{ \vspace{-\baselineskip}\vspace{-1.1mm} }
	\bibliographystyle{abbrv}
    \bibliography{SliceGAN_biblio}

\end{multicols}
\end{document}


\title{Generating 3D structures from a 2D slice with GAN-based
dimensionality expansion \\ \vspace{0.7cm} \LARGE{Supplementary Information}\vspace{0.7cm}}
    \author[1]{Steve Kench 
    \thanks{s.kench19@imperial.ac.uk}}
    
    \author[1]{Samuel J. Cooper \thanks{samuel.cooper@imperial.ac.uk}}
    
    \affil[1]{{\textit{\footnotesize Dyson School of Design Engineering, Imperial College London, London SW7 2DB}}}
    
    \maketitle
	
\section{Algorithm for anistropic microstructures}
\begin{algorithm}[H] 
	\begin{algorithmic}[1]
		\Require $w_1, w_2, w_3$ and $\theta$, the trainable parameters for three perpendicular discriminators and $G$ respectively; the number of discriminator iterations per generator iteration $n_D$; the batch sizes $m_D$ and $m_G$ for $D$ and $G$ backpropagation respectively; the gradient penalty coefficient, $\lambda$;  volume edge length, $l$; and Adam hyperparameters, $\alpha, \beta1, \beta2 $.
		\While {$ \theta $ has not converged }
		\State Discriminator training:
		\For {$t=0,...,n_{{D}}$ }
		\For {$i = 1, ..., m_D$}
		\State Sample a latent vector from a normal distribution $ \textbf{z} \sim p(\textbf{z})  $ 
		\State $\textbf{f} \leftarrow G_{\theta}(\textbf{z})$ generate a 3D volume
		\For {$a = 1,2,3$}
		\For {$d = 1,...,l$}
		\State $\textbf{f}_{\textbf{s}} \leftarrow$  2D slice of $\textbf{f}$ at depth $d$ along axis $a$
		\State Sample an $l \times l $ image from the real dataset associated with axis $a$ $ \textbf{r}^a \sim \mathbf{P}_\textbf{r}^a$ 
		\State Sample a random number $\epsilon \sim \mathbb{U}[0,1]$
		\State $\textbf{k} \leftarrow \epsilon \textbf{f}_{\textbf{s}} + (1-\epsilon)\textbf{r}^a$
		\State $L_D^{(a,d)} \leftarrow D_{w_a}(\textbf{f}_\textbf{s}) - D_{w_a}(\textbf{r}) + \lambda(\Vert \nabla_{\textbf{k}} D_{w_a}((\textbf{k}) \Vert_2-1)^2$
		\EndFor
		\State $w_a \leftarrow \text{Adam}(\nabla_w \frac{1}{m_D}\sum_{d = 1}^{l}
		L_D^{(a,d)})$
		\EndFor
		\EndFor
		\EndFor
		\State Generator Training:
		\For {$j = 1,...,m_G$}
		\State Sample a latent vector from a normal distribution $ \textbf{z} \sim p(\textbf{z})  $ 
		\State $\textbf{f} \leftarrow G_{\theta}(\textbf{z})$ generate a 3D volume
		\For {$a = 1,2,3$}
		\For {$d = 1,...,l$}
		\State $\textbf{f}_{\textbf{s}} \leftarrow$  2D slice of $\textbf{f}$ at depth $d$ along axis $a$
		\State $L_G^{(a,d)} \leftarrow -D_{w_a}(\textbf{f}_{\textbf{s}})$
		\EndFor
		\State $\theta \leftarrow \text{Adam}(\nabla_\theta \frac{1}{m_G}\sum_{d = 1}^{l} L_G^{(a,d)})$
		\EndFor
		
		\EndFor
		
		\EndWhile
		
		\caption{SliceGAN algorithm for anisotropic materials. Note the key change compared to isotropic is the use of different real data and a different discriminator for each axis.}
		\label{alg:SliceGanAlgo}
	\end{algorithmic}
\end{algorithm}
\vspace{0.5cm}
\section{Non-uniform density where $s$ is not divisible by $k$}
\label{sup:infoden}
\begin{figure}[H]
\includegraphics[width=\textwidth,]{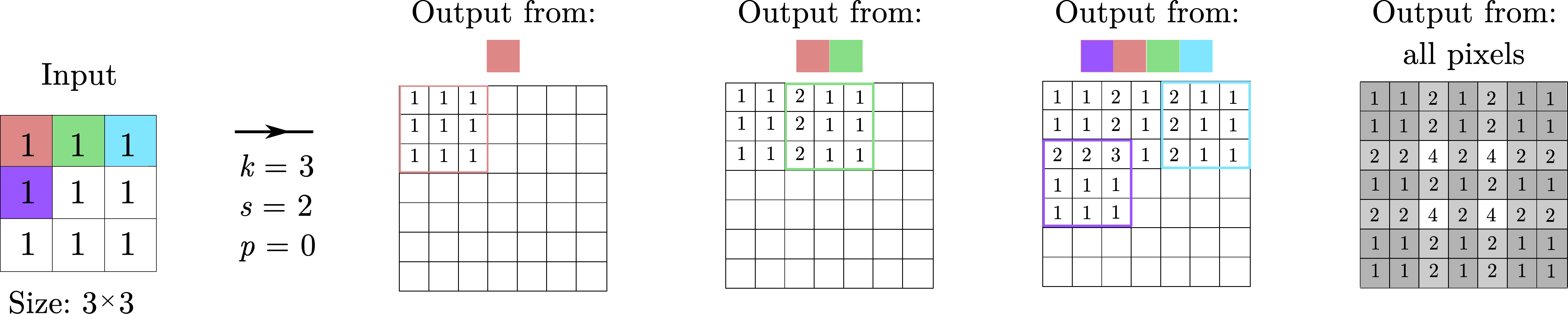}]
\caption{A transpose convolution example where kernel size ($k$) is not divisible by stride ($s$), resulting in a checkerboard, non-uniform information density. The fundamental cause of this behaviour is clear in the output from all four colored pixels. The cyan box shows how the convolution skips the central column, which is left with a low information density of 1 compared to its neighbours. This often results in unrealistic periodic features in generated images.}

\end{figure}
\section{Information density for a resize-convolution network}
\label{Sup:reco}
Figure S2 shows the edge information gradients associated with resize convolutions when using padding of 1 for each conv. Whilst it's possible to use no padding, which does yield uniform information gradient, this requires a big input vector (min $8 \times 8 \times 8$) to ensure the spatial dimensions don't collapse in the first layer, where normally padding would be used to enlarge the output. Large $\mathbf{z}$ significantly limits the number of filters that can be used in the early layers due to computational expense (for a 12GB GPU, the first layer can have $\approx 64$ filters, whilst a transpose convolution can have 512). This leads to poorer quality microstructures.
\begin{figure}[H]
	\begin{centering}
	\includegraphics[width=0.8\textwidth]{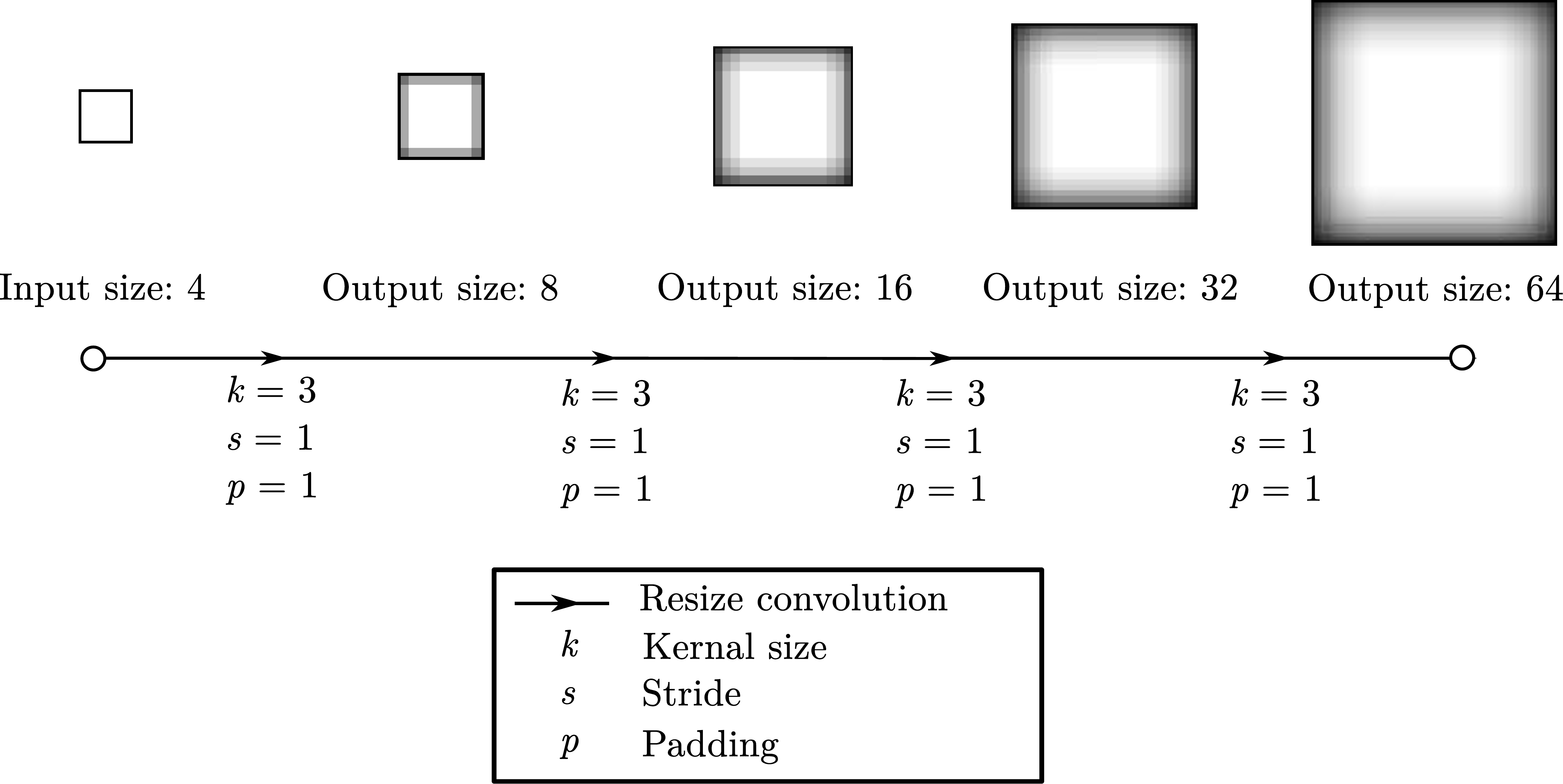}]
	\caption{A typical resize-convolution generator using zero-padding and upscaling factor of 2 between each convolution, showing the low information density near the image edges.}
	\end{centering}
\end{figure}
\section{Information overlap during post-training generation}
\label{Sup:RGB}
\begin{figure}[H]
	\includegraphics[width=\textwidth,]{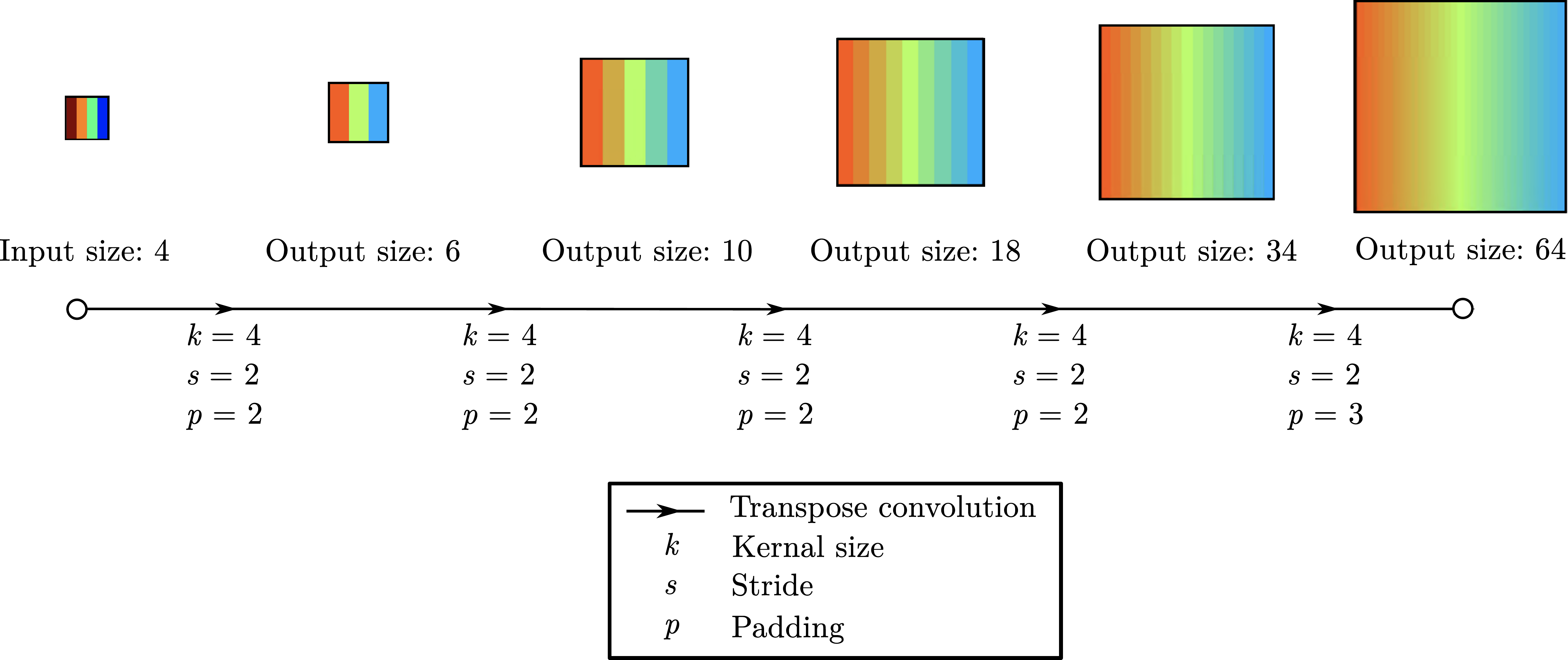}]
	\caption{A transpose convolution example using a $4\times4$ coloured input, showing the overlap of information during generation. The dark red input stripe spreads over to exactly half way across the image, with a linearly decreasing contribution to the output. Note that each of the RGB channels is calculated individually then concatenated and normalised.}
	
\end{figure}
\section{Periodicity of kernel element combinations}
\label{Sup:RGB_kernels}
\begin{figure}[H]
	\includegraphics[width=\textwidth,]{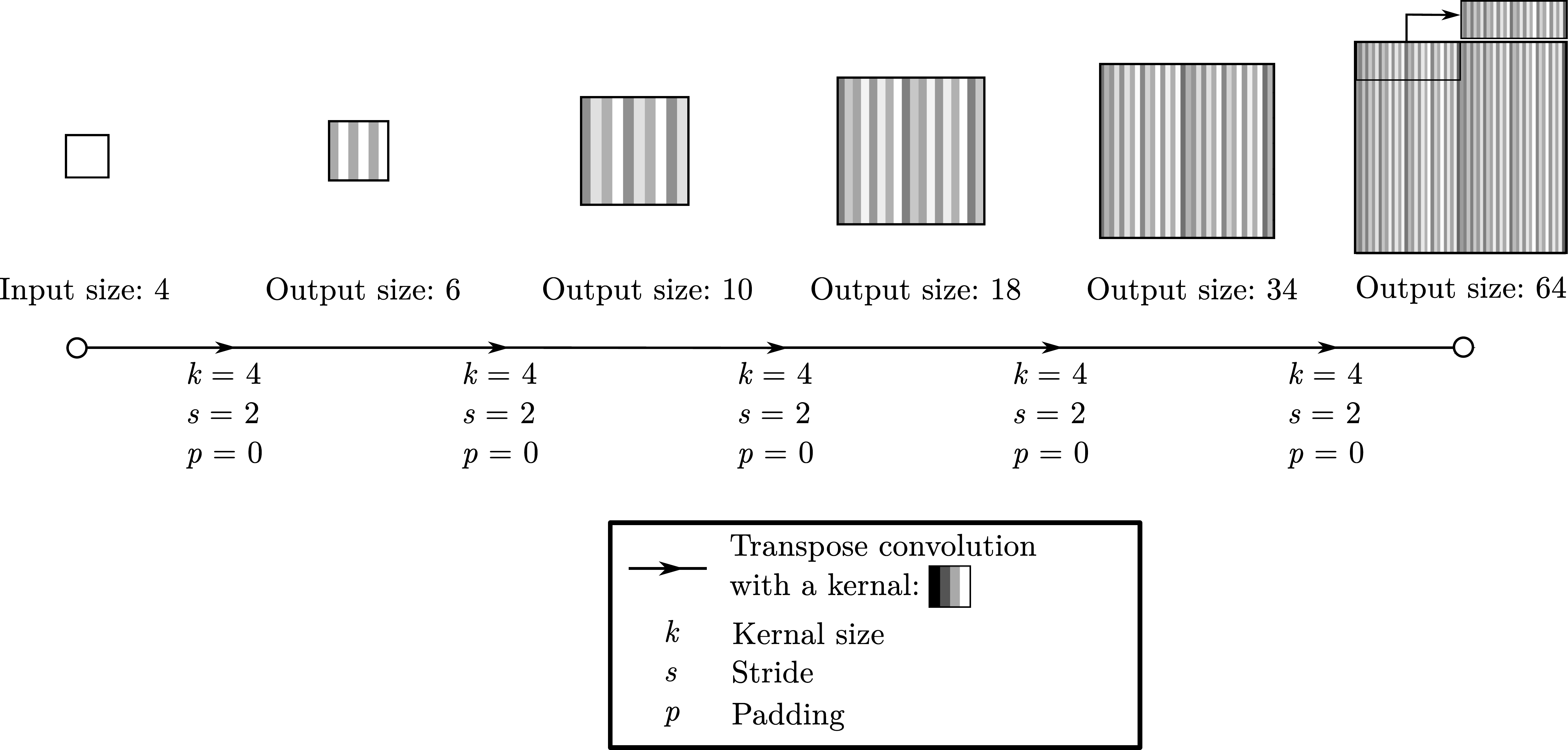}]
	\caption{A transpose convolution example using a striped kernel with columns of value 1, 2, 3, and 4, showing the periodic behaviour of the generator. In the final output, only slices that are generated from the exact same sequence of kernel elements are the same darkness - this occurs for every 32$^{\text{nd}}$ slice, whilst all other pairs are unique.}
	
\end{figure}
\section{One-hot coded transformation}
\label{Sup:OHE}
\begin{figure}[H]
	\centering
	\includegraphics[width=\textwidth,]{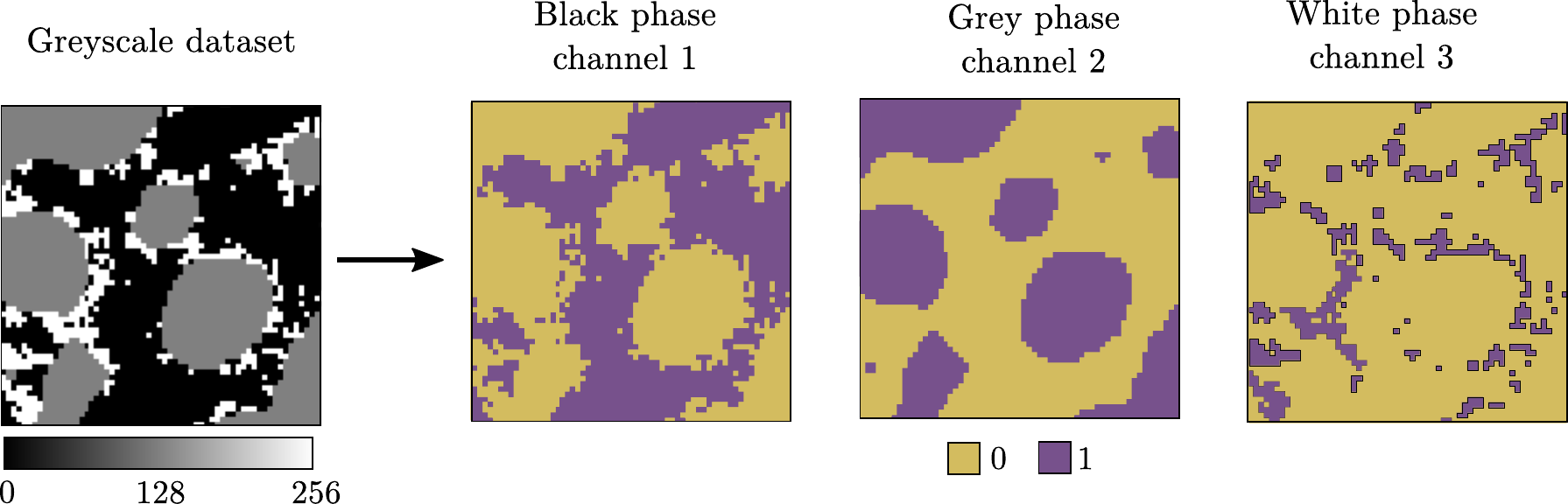}
	\caption{One-hot coding transformation starting from a segmented 8-bit grayscale image where each value corresponds to a particular phase, though their order is typically random. Whilst training image pixels only take binary values, generated samples instead have a probability distribution between 0 and 1 due to the $softmax$ function in the final layer.}

\end{figure}
\section{Metric comparison of real and synthetic datasets}
\begin{figure}[H]
	\centering
	\includegraphics[width=\textwidth,]{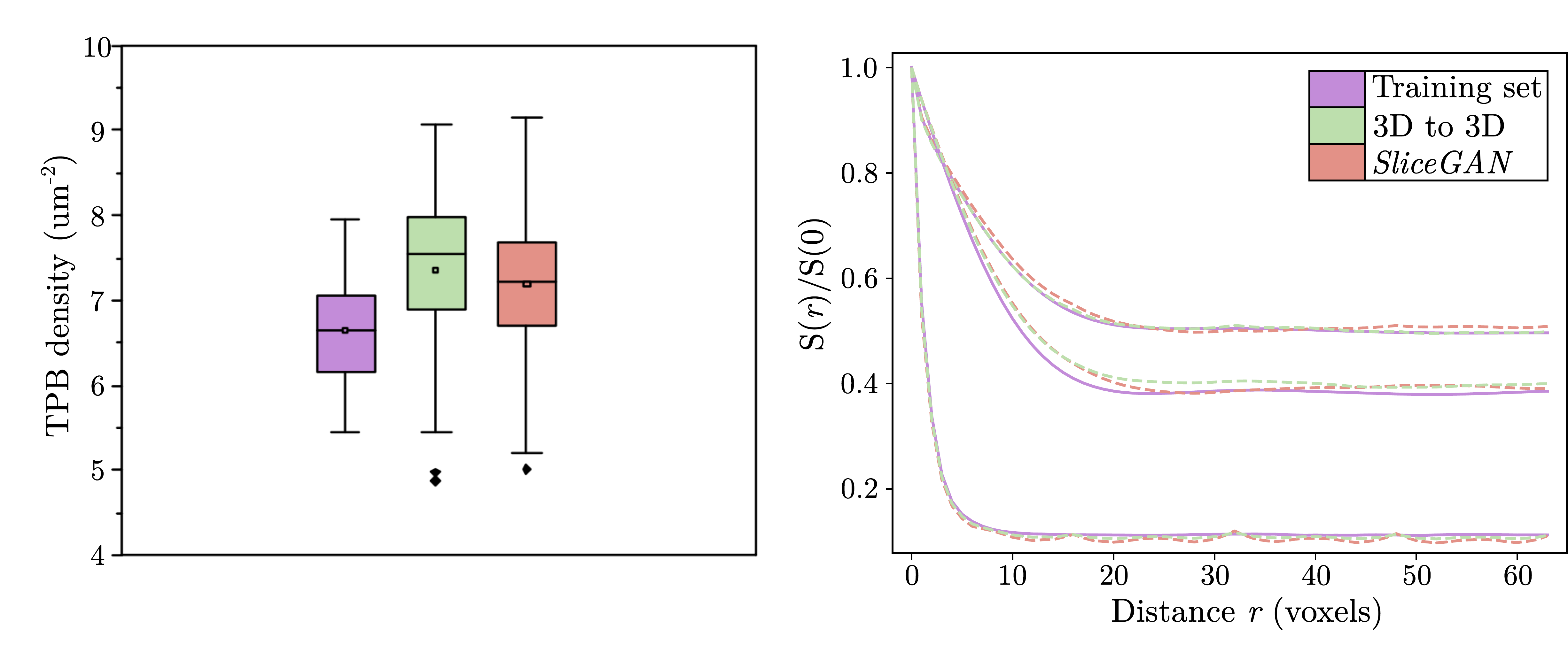}
	\caption{Triple phase boundary (TPB) density and two-point correlation (S($r$)) functions show the similarity between real and synthetic microstructures. TPB density is a measure of how many voxel vertices are shared by all three phases for a given volume. The median and distributions match well for the 100 samples used in this test. The two-point correlation function describes the probability of finding the same phase at voxels separated by distance $r$. In this case, we normalise this function against volume fraction, S(0), for ease of plotting. The values shown were measured using a single 250 edge length cube (corresponding to the full training data size). Again, the phase distributions match well.}
	
\end{figure}